# Joint Intensity-Gradient Guided Generative Modeling for Colorization

Kai Hong, Jin Li, Wanyun Li, Cailian Yang, Minghui Zhang, Yuhao Wang, *Senior Member, IEEE*, Qiegen Liu, *Senior Member, IEEE*

*Abstract*—This paper proposes an iterative generative model for solving the automatic colorization problem. Although previous researches have shown the capability to generate plausible color, the edge color overflow and the requirement of the reference images still exist. The starting point of the unsupervised learning in this study is the observation that the gradient map possesses latent information of the image. Therefore, the inference process of the generative modeling is conducted in joint intensity-gradient domain. Specifically, a set of intensity-gradient formed high-dimensional tensors, as the network input, are used to train a powerful noise conditional score network at the training phase. Furthermore, the joint intensity-gradient constraint in data-fidelity term is proposed to limit the degree of freedom within generative model at the iterative colorization stage, and it is conducive to edge-preserving. Extensive experiments demonstrated that the system outperformed state-of-the-art methods whether in quantitative comparisons or user study.

## I. INTRODUCTION

Colorization is the task of assigning color to each pixel of the target grayscale image. It has abundant applications in a variety of computer vision tasks, such as image compression [1], outline and cartoon creation colorization [2], [3], and remote sensing images colorization [4], [5]. Practically, coloring grayscale image is hard to address because the colorization results could be multiple viable for a single grey-level image [6]. Different colors can be assigned to the gray pixels of an input image, making the problem ill-posed and inherently ambiguous (e.g., the color of an apple can be red, green or irrational purple). Besides, the colorization problem also inherits the challenges of image enhancement, such as changes in illumination, variations in viewpoints, and occlusions. Therefore, colorization is still confronting challenges and waiting to be explored in image processing and computer vision.

At present, many methods have been developed to tackle the issues of colorization, which can be roughly divided into two categories: Interactive methods [7]-[14] and automatic methods [15]-[26]. Generally speaking, some of interactive methods require users to provide considerable scribbles, then determine the colors of the remaining pixels automatically [8]. The results of the colorization can differ greatly depending upon how the color scribbles are chosen. Furthermore, because the system purely relies on user inputs for colors, even regions with little color uncertainty need to be specified. Hence, the scribbling process is time-consuming and requires expertise. Example-based colorization is another reliable interactive method, which automatically searches for the most similar image patch/pixel for colorization [9]. The key point to the success of example-based colorization is how similar the reference image to the grayscale input. Regretfully, finding a suitable reference image becomes an obstacle for users.

To address these limitations, researchers have also explored deep learning in supervised or unsupervised manners for automatic colorization. These methods have achieved fantastic success in large-scale data. In addition, some methods introduced auxiliary tasks like image segmentation and fusion [20], auxiliary input text [10], [11], etc. to improve the colorization performance. Unsupervised learning is widely considered as the future direction of the colorization problem. It has the characteristics of obtaining implicit prior information and generating natural coloring results with diversity [27]. For instance, Vitoria *et al.* [22] proposed an automatic end-to-end adversarial approach that joints the advantages of generative adversarial networks (GANs) with semantic class distribution learning.

As mentioned above, most existing automatic colorization methods are usually optimized for intensity-level information. Nevertheless, since the grayscale image lacks detailed colorful characteristics, the results often contain incorrect colors and obvious artifacts. Currently, some methods try to extract semantic-level information to improve colorization performance [20], [28]. But this kind of high-level image understanding is difficult to be captured and may result in unsuitable effects. Therefore, we attempt to extract the features in gradient domain of the grayscale image as "guidance" information.

On the one hand, recent study has shown that the human visual system (HVS) cannot perceive absolute pixel values but relies on the local contrast ratio [29], which is directly related to the gradient information. Gradient domain information has a wide range of applications in many works of image processing. It has achieved great success in image enhancement [30]-[32], image restoration [33], [34], and image fusion [35], etc. Learning the gradient distribution of natural images can improve the quality of subjective images. Furthermore, equalizing histogram in the gradient domain can also achieve the effects of image deblurring and smoothing [36]. On the other hand, as one of the basic information of the image, conducting unsupervised learning in gradient domain can expand the dimensionality of the data density distribution. Especially, this strategy adds more prosperous prior information on the learning and inference

This work was supported by National Natural Science Foundation of China (61871206, 61601450).

K. Hong, J. Li, W. Li, C. Yang, M. Zhang, Y. Wang and Q. Liu are with the Department of Electronic Information Engineering, Nanchang University, Nanchang 330031, China. ({hongkai, lijin, liwanyun, yangcailian}@email.ncu.edu.cn, {zhangminghui, wangyuhao, liuqiegen}@ncu.edu.cn).

processes, which adapts to the characteristics of the unsupervised learning process. For instance, Pan *et al.* [37] proposed a simple yet effective $L0$-regularized prior based on intensity and gradient for text image deblurring. They pointed out that this image prior was based on distinctive properties of text images and did not require any heuristic edge selection methods which were critical to state-of-the-art edge-based deblurring methods.

In this work, we study how to effectively employ generative model in joint intensity-gradient domain to improve the performance of automatic colorization and the visual perception of colorized images. As far as we know, gradient information has already been exploited in supervised learning manner for colorization [38], while only enforced in loss function. The application of prior information in gradient domain to unsupervised automatic colorization is still a blank area worth exploring.

This study focuses on the field of automatic image colorization and presents a novel Joint intensity-gradient domain guided Generative Model for satisfactory colorization, coined JGM. In this work, we approach this task from the generative network of noise conditional score networks [39]. It uses Langevin dynamics to generate the gradients of data distribution. The colorization problem can be viewed as an image-generative problem enforced with linear constraints to generate a sound result. The rationality of the underlying idea in JGM is listed in Fig. 1. In Fig. 1(b), the generative model in joint intensity-gradient field achieves a more colorful result than only in intensity domain. Moreover, by enforcing linear constraints in both intensity and gradient domains, the degree of freedom of the generative model is better exploited, thus the colorization result in Fig. 1(c) exhibits more natural effects.

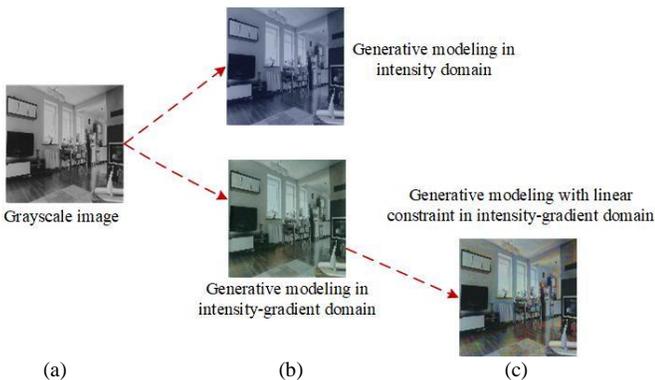

**Fig. 1.** Visualization of the colorization results with different generative modelings. (a) The reference grayscale image. (b) The top line is the colorization result with generative modeling in intensity domain, and the bottom line is with the generative modeling in joint intensity-gradient domain. (c) The colorization result of the proposed JGM. Particularly, the generative modeling of JGM is conducted in 9-channel (intensity-gradient) domain, largely reduces the color ambiguity in intensity and attains a more natural and realistic result.

The major contributions of this paper are:
- **Iterative generative modeling in joint intensity-gradient domain:** A novel automatic colorization via score-based generative modeling is used for exploring the prior information in joint intensity-gradient domain. Learning prior knowledge in redundant and high-dimensional subspace paves the way for producing more chances to attain diversity and possible colorization.
- **Enforcing linear constraints in image intensity-gradient field:** Aid by the linearity property of the gradient operator, linear data-fidelity constraint is also enforced in gradient domain. The constraints in joint intensity-gradient domain leverage the visual appearance of results, leading to preferable colorization in high-dimensional space.

The rest of this paper is presented as follows. Section II briefly describes some relevant works on colorization, gradient-based processing and denoising score matching (DSM) [40]. In section III, we elaborate on the formulation of the proposed method. Section IV presents the colorization performance of the present JGM, including the experimental setup, colorization diversity, and comparisons with the state-of-the-art methods. In section V, some variants of the standard JGM are discussed. Finally, Section VI summaries our contributions, and future works are given for the end.

## II. RELATED WORK

### A. Colorization

Colorization is an ill-posed problem due to the large degree of freedom during the assignment of color information. Mathematically, the common approach in the literature to solve the problem relies on the following linear observation model:
$$y = Fx + n \qquad (1)$$
where $y \in \mathbb{R}^N$ and $x \in \mathbb{R}^N$ denote the gray-level image and the original color image, respectively. $n \in \mathbb{R}^N$ denotes the additive white Gaussian noise. $F \in \mathbb{R}^{N \times N}$ denotes the degradation matrix for graying $x$. With regard to the forward operator, some researchers assume the lightness channel $L$ in CIE Lab color space [21] or the luminance channel $Y$ in YUV color space as the grayscale observation [15]. In this work, the grayscale image is represented by $Fx = (x_r + x_g + x_b)/3$.

Existing work on colorization fall into two major categories: interactive methods [7]-[14] and automatic methods [15]-[26]. Interactive colorization methods depend on user-provided guidances, such as color scribbles [8], [10], [11] or reference images [9], [12]-[14], which are the key point of obtaining the satisfactory result of colorization. Levin *et al.* [8] proposed a colorization method based upon an optimization problem. Li *et al.* [13] proposed a novel cross-scale texture matching method to improve the robustness and quality of the examples-based colorization results. These methods can achieve impressive results but often require intensive user interaction, as each different color region must be explicitly indicated by the user. Simultaneously, finding a suitable reference image is also a severe obstacle.

Deep learning for automatic approaches has shown the ability to capture more intricate color properties on a huge amount of grayscale and color image pairs [15]-[26]. Cheng *et al.* [17] proposed the first deep neural network model for fully automatic image colorization. It used adaptive image clustering technique to incorporate the global image information through joint bilateral filtering. Zhang *et al.* [21] utilized the self-supervised classification network in different color spaces to achieve automatic colorization. Besides, generative model [41], [42] is an essentially unsupervised tool to describe the dataset. ChromaGAN [22] exploited the semantic understand-ing of the real scenes via GAN. Messaoud *et al*. [23] developed a conditional random field-based variational autoencoder formulation to achieve diversity. The generative model is capable to learn probability distributions over different color spaces of data and has been widely used for many tasks.

### B. Gradient-based Image Processing

Since the HVS is more sensitive to the gradient map than to the absolute intensity of an image, gradient domain information is widely used in image process to attain satisfactory visual effect [29]. Gong *et al*. [30] proposed a method called "image naturalization", learning the specified gradient distribution and remapping the image to match it. Cho *et al*. [34] proposed an iterative distribution reweighting (IDR) method, which made gradient distribution of the reconstructed images similar to the reference distribution. Besides, gradient histogram equalization of the image is another widely used image enhancement method. Wang *et al*. [33] suggested tackling image enhancement and restoration problems in the gradient domain instead of the traditional intensity domain. Petrovic *et al*. [35] represented the input image as a gradient map and then merged it to generate a new fused image without losing information or introducing distortion.

Although the usage of gradient information is wide in various image processing applications, the research on using the gradient in automatic image colorization is few. It is predictable that introducing generative modeling in gradient domain is a promising direction. For instance, the specified distribution of gradient domain is independent of image content [30], [38], so that the generative model can adapt to a wider range of images with different objects.

### C. Prior Learning: From DAE to DSM

Unlike the supervised methods that rely on a carefully designed prior, the unsupervised methods learn general and explicated prior functions to infer the missing details from the observed image. By taking advantage of the high nonlinearity and capacity of neural networks, autoencoder (AE) has become one of the most important unsupervised learning methods for the task of representation learning [42], [43]. Denoising autoencoder (DAE) is a simple modification of the classical autoencoder that not to reconstruct their input itself, but to denoise an artificially corrupted version. It essentially acts as a regularized criterion for learning to capture useful structure from the input data. [44]-[46].

Bigdeli *et al*. [47] used the magnitude of the autoencoder error as a prior (DAEP) for image recovery and the DAE $A_\sigma$ is trained to minimize the expectation over all input images $x$:

$$L_{DAE}(A) = E_{x,\eta}[\|A_\sigma(x+\eta) - x\|^2] \qquad (2)$$

where the output $A_\sigma(x)$ is trained by adding artificial Gaussian noise $\eta$ with standard deviation $\sigma$.

According to [48], the network output $A_\sigma(x)$ is related to the true data density $p(x)$ as follows:

$$A_\sigma(x) = x + \nabla \log \int g_\sigma(\eta) p(x+\eta) d\eta \qquad (3)$$

where $g_\sigma(\eta)$ represents a local Gaussian kernel with standard deviation $\sigma$.

To link to the score matching, Vincent proposed the DSM, which is equivalent to DAE [40]. Furthermore, Block *et al*. shed new light on what the DAE learns from a distribution, showing that optimizing DAE loss is equal to optimizing DSM loss [48] (e.g., let $p$ be a differentiable density). The DSM loss is as follows:

$$L_{DSM}(s) = E_{p_\sigma}[\|s(x) - \nabla \log p_\sigma(x)\|^2] \qquad (4)$$

with

$$s(x) = \frac{A(x) - x}{\sigma^2} \qquad (5)$$

The DAE loss and the DSM loss are equivalent to a term that does not depend on $A$ or $s$. It is seen that DSM optimizes data distribution more directly.

## III. PROPOSED JGM MODEL

### A. Motivation of JGM

The present JGM is motivated by our previous work in iterative generative model (iGM) [50]. Specifically, unlike the traditional generative models that often encode the input to be a latent variable in lower-dimensional spaces (e.g., CVAE [49], [25]), the generative model of iGM is exploited in multi-color spaces (e.g., RGB, YCbCr) jointly. It benefits from the high-dimensional inference process of generative modeling. It should be mentioned that the high-dimensional learning schemes have been studied in several recent works [50], [52]. They pointed out that the prior information learned from high-dimensional tensors is more effective than the information obtained from low-dimensional objects.

JGM inherits the great potential of iGM that conducting generative modeling in high-dimensional space. In iGM, as the multi-color space information cannot be captured directly in grayscale image, it needs to conduct generative modeling progressively. Hence, computational cost will increase largely as more multi-color spaces are utilized. To alleviate the issue, finding a high-dimensional variable that can be extracted from the grayscale image directly is urgent. Motivated by the observation that image gradient is an effective tool to describe the inherent property in the image itself, JGM exploits high-dimensional prior information in joint intensity-gradient domain.

To facilitate the exposition, the visual comparison of CVAE [25], iGM [50] and JGM is shown in Fig. 2. In Fig. 2(a), CVAE learns a multi-modal conditional model $P(z|y)$, between the low-dimensional embedding mapping $z$ and the grayscale image $y$. Under the mixture of Gaussians is used as the input [51], the conditional probability distribution of target vectors is modeled by mixture density networks. The one-to-many mapping allows the target vectors to take multiple values for the same input vector, which provides diversity. It can be seen in Fig. 2(b) that the iGM expands the freedom of color by high-dimensional tensor and then adopts the intermediate to obtain linearly autocorrelative constraint that guides the direction of the color more precisely.

The diagram of JGM is depicted in Fig. 2(c). As seen, JGM consists of prior learning and iterative colorization stages. At the prior learning stage, a set of high-dimensional tenors is used to learn the prior information from the gradient of data distri-bution in the joint intensity-gradient domain. At the iterative colorization stage, under the linear constraint of the intensity-gradient domain, we utilize Langevin dynamics to generate more accurate colorizations.

### B. JGM: Prior Learning in Intensity-Gradient Domain

In recent years, likelihood-based models (e.g., variational autoencoder (VAE), AE) and GANs have achieved great success, but they always have some intrinsic limitations about forming good prior knowledge to inform parameter learning [15], [22], [25]. Lately, Song *et al.* [39] introduced a new generative model named noise conditional score networks. The model estimates the gradients of data distribution by DSM and then produces samples progressively via Langevin dynamics. The framework allows flexible model architectures, avoids using adversarial methods and does not require sampling during training. Besides, it also provides a learning objective that can be used for principled model comparisons.

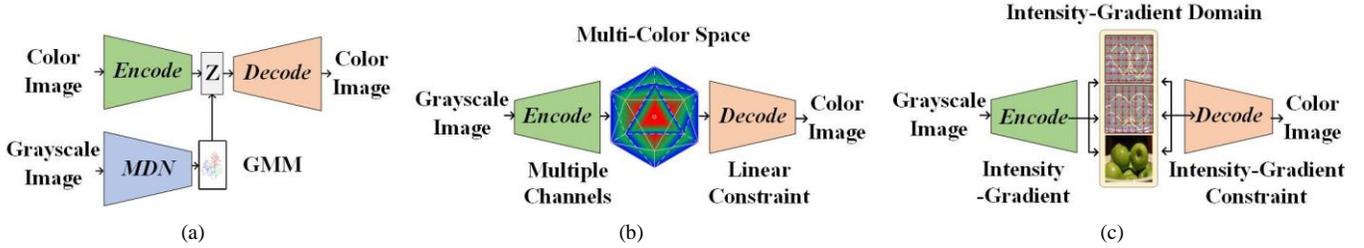

**Fig. 2.** Visual comparison of CVAE (a), iGM (b) and JGM (c). Rather than the CVAE that optimally encode the compressible latent space to achieve the colorization goal, both iGM and JGM utilize the generative modeling in high-dimensional space to optimize the colorization target. Particularly by taking advantage of the joint intensity-gradient field, the proposed JGM learns prior information and iteratively approaches to color image.

Through score functions, the score-based generative model explores a new principle, which progressively denoises an initial white noise into an image with the output of score-based network. It directly trains a score network $s_\theta(x)$ to estimate the gradient of data prior $\nabla_x \log p_{data}(x)$ instead of data prior $p_{data}(x)$. The score function is learned via denoising score matching [53], [54]. In this process, the key insight is to perturb the data using multiple noise levels to help the score network capture both coarse and fine-grained image features. The noise distribution is chosen to be $p_\sigma(\tilde{x}|x) = N(\tilde{x}|x, \sigma^2 I)$; therefore $\nabla_{\tilde{x}} \log p_\sigma(\tilde{x}|x) = -(\tilde{x}-x)/\sigma^2$. The score-based generative model aims to train a conditional score network to estimate the scores of all perturbed data distributions jointly, i.e., under a sequence of noise levels $\{\sigma_i\}_{i=1}^L$. More specifically, for a given $\sigma$, it estimates the score function of each $p_\sigma(x)$ by training a single neural network $s_\theta(x, \sigma)$ with the DSM loss:

$$\ell(\theta; \sigma) \triangleq \frac{1}{2} E_{p_{data}(x)}[\|s_\theta(x,\sigma) - \nabla_x \log p_\sigma(x)\|_2^2]$$
$$= \frac{1}{2} E_{p_{data}(x)} E_{p_\sigma(\tilde{x}|x)} \|s_\theta(\tilde{x},\sigma) - \nabla_{\tilde{x}} \log p_\sigma(\tilde{x}|x)\|_2^2 \quad (6)$$
$$= \frac{1}{2} E_{p_{data}(x)} E_{p_\sigma(\tilde{x}|x)} \left\|s_\theta(\tilde{x},\sigma) + \frac{\tilde{x}-x}{\sigma^2}\right\|_2^2$$

Then, Eq (6) is combined for all $\sigma \in \{\sigma_i\}_{i=1}^L$ to get one unified objective:

$$L(\theta; \{\sigma_i\}_{i=1}^L) \triangleq \frac{1}{L} \sum_{i=1}^L \lambda(\sigma_i) \ell(\theta; \sigma_i) \quad (7)$$

where $\lambda(\sigma_i) > 0$ is a coefficient function depending on $\sigma_i$. As a conical combination of DSM objectives, one can imagine that Eq. (7) achieves the minimum value if and only if $s_{\theta^*}(x, \sigma_i) = \nabla_x \log p_{\sigma_i}(x)$ for all $i \in \{1, 2, \cdots, L\}$. Indeed, perturbing the data with various noise levels and training a single conditional score network by simultaneously using the estimated scores corresponding to all the levels largely improve the performance of score-based generative model.

In this work, we introduce the gradient information of image into the automatic colorization. In short, we estimate the gradients of the RGB data distribution and its gradient distribution with DSM. The score-based network is conducted in intensity-gradient domain. Specifically, denoting the target color image containing the $R$, $G$ and $B$ components to be $x = (x_r, x_g, x_b)$, then, its gradients in vertical and horizontal directions are as follows:

$$\nabla_1 x = (\nabla_1 x_r, \nabla_1 x_g, \nabla_1 x_b)$$
$$\nabla_2 x = (\nabla_2 x_r, \nabla_2 x_g, \nabla_2 x_b) \quad (8)$$

Accordingly, for every continuously differentiable probability density $p(X)$ of $X = [x, \nabla x] = [x, \nabla_1 x, \nabla_2 x]$, we call $\nabla_X \log p(X)$ as its score function. At the training phase, we implement 9-channel (9C) high-dimensional tensor $X$ as input of the network to learn the gradients of the data distribution $\nabla_X \log p_{\sigma_i}(X)$ by training a single neural network $s_\theta(X, \sigma)$ with the following loss:

$$\frac{1}{2L} \sum_{i=1}^L E_{P_{data}(X)} E_{p_{\sigma_i}(\tilde{X}|X)} \lambda(\sigma_i) \left\|s_\theta(\tilde{X},\sigma_i) + \frac{\tilde{X}-X}{\sigma_i^2}\right\|_2^2 \quad (9)$$

The essence of stacking to be $X$ is to obtain much more data information at high-dimensional manifold and high-density regions, thus avoiding some difficulties in score estimation of DSM [39], [40]. Subsequently, the JGM is trained over data $X$ in high-dimensional space as network input and the parameterized $s_\theta(X, \sigma)$ is obtained. The visualization of the prior learning stage is depicted in the top region of Fig. 3. Generally speaking, we try to learn existing information on a higher-dimensional basis by combining the image domain and the gradient domain as channels through channel concatenation and setting them as network inputs for joint training.

### C. JGM: Iterative Colorization in Intensity-Gradient Field

The entire colorization procedure mainly involves three ingredients in the testing phase: Annealed Langevin dynamics, intensity-gradient constraints, and linear minimization.

Annealed Langevin dynamics is introduced as a sampling approach. Once the score function is known, the Langevin dynamics can be applied to sample from the corresponding distribution. In the iterative colorization process, samples of each disturbance noise level are used as the initial input of next noise level until reaching the smallest one which provides samples for the network to generate final colorization result gradually. Specifically, given a step size $\alpha > 0$, the total number of iterations $T$, and an initial sample $X_t$ from any prior distribution, Langevin dynamics iteratively evaluate the following:

$$X_{t+1} \leftarrow X_t + \alpha \nabla_X \log p(X_t) + \sqrt{\alpha} z_t \quad (10)$$

where $\forall t : z_t \sim N(0, I)$.

Supposing we have a neural network $s_\theta(X)$ parameterized by $\theta$, and it has been trained such that $s_\theta(X, \sigma_i) \approx \nabla_X \log p(X, \sigma_i)$ for all $i \in \{1, 2, \cdots, L\}$. By decreasing the $\alpha_i$-value (accordingly $\sigma_i$-value), we can approximately generate samples from $p(X)$ using annealed Langevin dynamics by replacing $\nabla_X \log p_{\sigma_i}(X_t)$ with $s_\theta(X_t, \sigma_i)$ iteratively, i.e.,

$$X_{t+1} \leftarrow X_t + \frac{\alpha_i}{2} s_\theta(X_t, \sigma_i) + \sqrt{\alpha_i} z_t \quad (11)$$

Because of the great randomness of generative model, a new constraint called data-consistency flow that imposes on the iteration process is proposed to make the colorization process more controllable and the colorization effect more natural. Thus, the modified Annealed Langevin dynamics combined with intensity-gradient data-consistency flow can be expressed as:

$$X_{t+1} \leftarrow X_t + \frac{\alpha_i}{2}[s_\theta(X_t, \sigma_i) - \lambda DC(X_t)] + \sqrt{\alpha_i} z_t \quad (12)$$

where the data-consistency term $DC(X)$ contains the linear data-fidelity constraints in intensity and gradient domains, respectively, i.e., $DC(X) = DC_1(X) + DC_2(X)$ and

$$\begin{aligned} DC_1(X) &= DC_1(x) = (Fx - y) \\ DC_2(X) &= DC_2(\nabla x) = (F\nabla x - \nabla y) \end{aligned} \quad (13)$$

The visualization of the iterative colorization stage is depicted in the bottom region of Fig. 3.

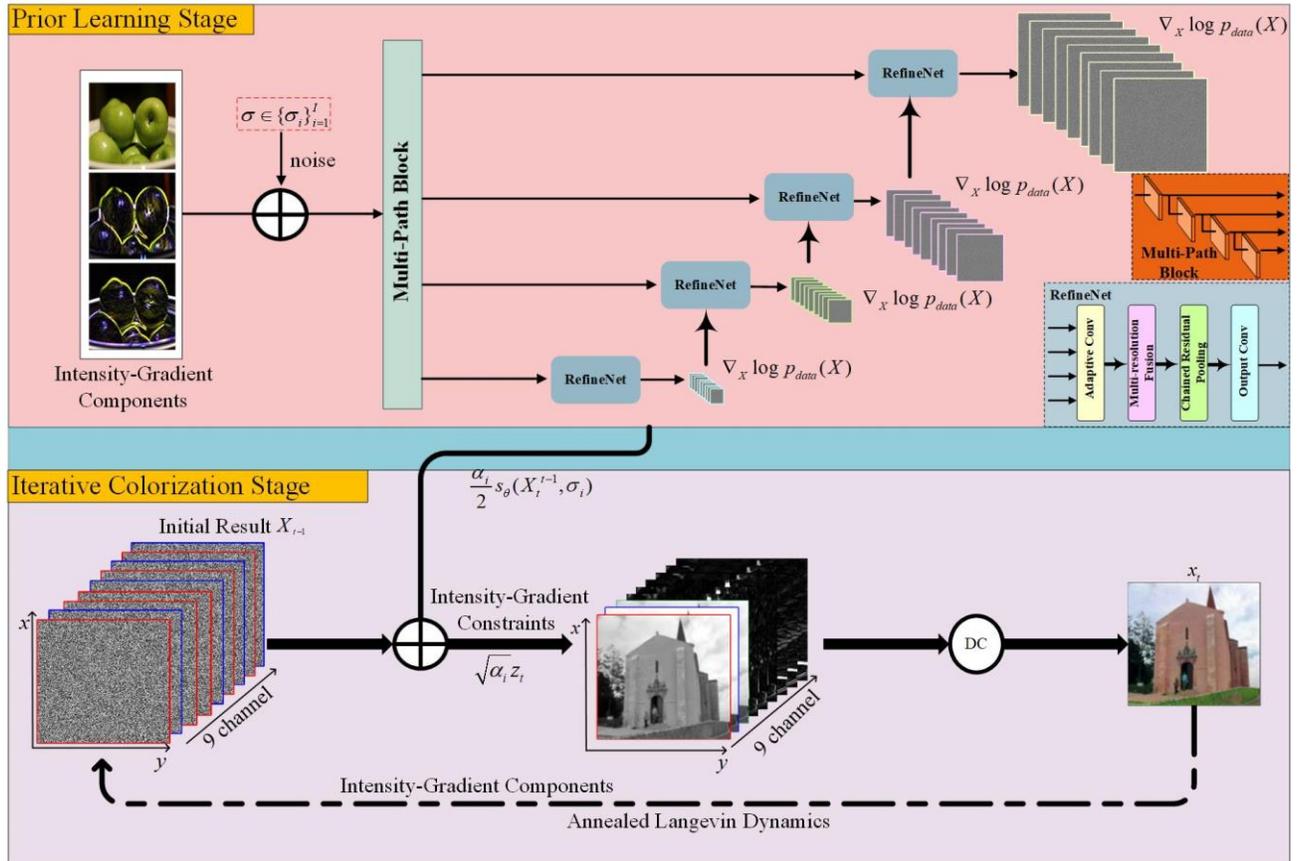

**Fig. 3.** The pipeline of the prior learning stage and the iterative colorization procedure of JGM. More specifically, the prior training stage learns the data distribution (including images domain and gradients domain) from the reference dataset, which acts as prior information for later colorization. The colorization stage generates samples from the high-dimensional noisy data distribution by annealed Langevin dynamics, under the given intensity-gradient data-consistency constraint. Here symbol "⊕" stands for the sum operator and "DC" stands for the data-consistency in intensity-gradient domain.

Finally, we apply the intermediate result to update the image $x_{t+1}$ and the gradient maps $[\nabla_1 x_{t+1}, \nabla_2 x_{t+1}]$. Intuitively, colorized image from joint intensity-gradient field can be attained by direct intensity-gradient integration with solving Poisson equation [55]. However, considering that solving the Poisson equation is an ill-posed problem and may introduce a large error, we sort to a weighting strategy. The approach is a linear minimization with gradient fidelity and image consistency. Obviously, the linear summation incorporates the local information in gradient domain and global information in intensity domain simultaneously. Denoting the final output as $x$, the mixed minimization is formulated as:

$$\min_{x}\{\|x - x_{t+1}\|^2 + \beta [\|\partial_1 x - \nabla_1 x_{t+1}\|^2 + \|\partial_2 x - \nabla_2 x_{t+1}\|^2]\} \quad (14)$$

where the first term is the color confidence to use the latest image to guide the image colorization. The second term is the gradient confidence that is close to the intermediate gradient results. $\beta$ is a parameter balancing the two-loss functions. Contrary to Poisson integration that relies purely on the computed gradients with a boundary condition, our iterative colorization finds a balance between reference image and computed gradients, which is especially useful when gradients are not integrable.

In summary, as explained in **Algorithm 1**, the whole colorization procedure consists of two-level loops: The outer loop handles $\nabla_X \log p_{\sigma_i}(X)$ to approximate $\nabla_X \log p_{data}(X)$ of the intensity and gradient domains with decreasing $\sigma_i$-value, while the inner loop decouples to be an alternating process of the updating estimated gradient of data prior $\nabla_X \log p_\sigma(X)$ and the least square scheme.

To intuit the procedure of annealed Langevin dynamics, we provide intermediate samples and the associate PSNR values in Fig. 4. As can be seen, the samples are generated from the high-dimensional noisy data distribution, evolving from pure random noise to colorful images.

| **Algorithm 1. JGM for Iterative Colorization** |
|---|
| **Training stage** |
| **Dataset**: Intensity-gradient domain dataset: $X = [x, \nabla x] = [x, \nabla_1 x, \nabla_2 x]$ |
| **Outputs**: Trained JGM $s_\theta(X, \sigma)$ |
| **Iterative colorization stage** |
| **Initialize**: $\sigma \in \{\sigma_i\}_{i=1}^L, \varepsilon, L$, and $x$ |
| For $i \leftarrow 1$ to $L$ do |
| $\quad \alpha_i = \varepsilon \cdot \sigma_i^2 / \sigma_L^2 \quad z_t \sim N(0, I)$ |
| $\quad$ For $t \leftarrow 1$ to $T$ do |
| $\quad\quad$ Stack $X_t = [x_t, \nabla x_t]$ and Sample $z_t \sim N(0, I)$ |
| $\quad\quad X_{t+1} \leftarrow X_t + \dfrac{\alpha_i}{2}[s_\theta(X_t, \sigma_i) - \lambda DC(X_t)] + \sqrt{\alpha_i} z_t$ |
| $\quad$ End for |
| $\quad x_{t+2} = \arg\min_x\{\|x - x_{t+1}\|^2 + \beta [\|\partial_1 x - \nabla_1 x_{t+1}\|^2 + \|\partial_2 x - \nabla_2 x_{t+1}\|^2]\}$ |
| End for |
| **Output**: $x_T$ |

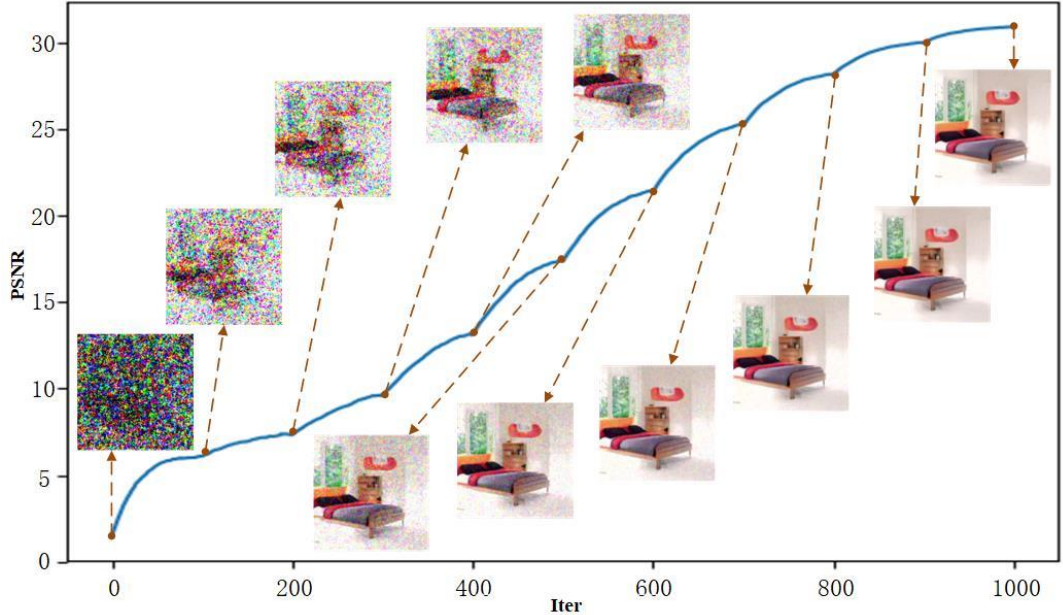

**Fig. 4.** Visualization of the intermediate colorization process with annealed Langevin dynamics. As the level of artificial noise becomes smaller, the colorization results tend to more natural color effects.

## IV. EXPERIMENTAL RESULTS

In this section, the proposed method is evaluated both quantitatively and qualitatively. First, the performance of JGM is compared against several state-of-the-art automatic colorization methods on different datasets and calculated their quantitative index values. A subjective user study is then performed to evaluate the results of different methods qualitatively. Finally, a set of colorization results are exhibited to validate the diversity ability of JGM. For convenient reproducibility, the source code of JGM is available at https://github.com/yqx7150/JGM.

### A. Experiment Setup

**Datasets**: We choose three datasets for experiments.

LSUN [56] (bedroom and church): LSUN is a large color image dataset. It contains around one million labeled images for each of 10 scene categories and 20 object categories.

Among them, we choose the indoor scene LSUN-bedroom dataset, which has enough samples (more than 3 million) and various colors to verify the effectiveness of JGM. On the other hand, in the outdoor scene LSUN-church dataset, the trees are almost all green and the sky is always blue. Most of the images fall into the same scene category.

COCO-stuff [57]: The Common Objects in COntext-stuff (COCO-stuff) is constructed by annotating the original COCO dataset, which originally annotated many complex daily things while neglecting stuff annotations. There are 164k images in COCO-stuff dataset that span over 172 categories, including 80 things, 91 stuff, and one unlabeled class.

ImageNet [58]: The largest clean image dataset for image classification in computer vision community, currently contains 3.2 million images with clear annotations and extremely rich types. ImageNet is widely used to validate the colorization performance to the most diverse images and various complex scenes.

In experiments of prior learning, we use 150,000 images randomly picked from each dataset. We reshape each image into 128×128 pixels as preprocessing. At the test stage, we randomly choose 100 images from the validation set for each dataset.

*Model Training*: The proposed JGM selects the UNet-type architectures with instance normalization and dilated convolutions as the network structure. Adam is chosen as an optimizer with a learning rate of 0.005 and halved every 5,000 iterations. Subsequently, the JGM model is trained for $1e5$ iterations with the batch size of 8 that takes around 20 hours. The model is performed with Pytorch interface on 2 NVIDIA Titan XP GPUs, 12GB RAM.

*Quality Metrics:* The main metrics used for comparisons among different methods in all reported experiments are the PSNR (Peak-Signal-to-Noise-Ratio), SSIM (Structural Similarity) and Naturalness [59]. Denoting $x$ and $\hat{x}$ to be the colorized image and ground-truth, the PSNR is defined as:

$$PSNR(x,\hat{x}) = 20\log_{10}\text{Max}(\hat{x})/\|x-\hat{x}\|_2 \quad (15)$$

The SSIM is defined as:

$$SSIM(x,\hat{x}) = \frac{(2x_x x_{\hat{x}} + c_1)(2\sigma_{x\hat{x}} + c_2)}{(\mu_x^2 + \mu_{\hat{x}}^2 + c_1)(\sigma_x^2 + \sigma_{\hat{x}}^2 + c_2)} \quad (16)$$

Besides, as the ultimate goal of image colorization is to obtain satisfactory results for the observer, we introduce the user study method to test reality and naturalness of results directly. This criterion will further test whether the algorithm can produce realistic, natural and reasonable coloring effects.

*Compared Methods*: The proposed method JGM is compared with four state-of-the-art methods, including Zhang *et al.* [21], MemoPainter [26], ChromaGAN [22] and iGM [50].

### B. Comparisons with State-of-the-arts

Quantitative comparisons are performed for the LSUN (i.e., bedroom and church) dataset, ImageNet dataset and COCO-stuff dataset to evaluate the compared algorithms. We randomly use 100 grayscales as input for each method and take the average score of PSNR and SSIM of the 100 final colorized images as the final result.

The results are tabulated in Table 1. For both PSNR and SSIM indexes, JGM achieves the highest value of 29.46 dB and 0.9365 on the LSUN-bedroom dataset. The PSNR result is almost 4 dB higher than the current state-of-the-art methods, such as ChromaGAN and iGM. For more complicated dataset COCO-stuff, JGM also achieves satisfactory performance.

Since colorization is regarded as a one-to-many task that multiple feasible colorized results may be given under the same grayscale input, JGM does not necessarily aim to restore the ground-truth color of the original image to pursue high PSNR and SSIM. Even so, JGM performs favorably against recent methods, highlighting its effectiveness.

TABLE I
COLORIZATION COMPARISON OF OUR SYSTEM TO STATE-OF-THE-ART TECHNIQUES, IN TERMS OF PSNR AND SSIM PERFORMANCE.

| Algorithms | LSUN-church | LSUN-bedroom | COCO-stuff |
|---|---|---|---|
| Zhang *et al.* | 23.65/0.9228 | 20.89/0.8946 | 20.21/0.8844 |
| MemoPainter | 22.85/0.9034 | 23.08/0.9004 | 20.47/0.8492 |
| ChromaGAN | **24.63**/0.9106 | 24.16/0.8899 | 19.85/0.8101 |
| iGM-6C | 23.12/0.9251 | 24.83/0.9221 | **21.95/0.8871** |
| JGM | 23.76/0.8931 | **29.46/0.9365** | 20.72/0.7254 |

The colorization results of different approaches are shown in Fig. 5 for the qualitative comparison. Although most of the methods can achieve basic colorization, the deficiency of unclear color range and low saturation still exists. It can be found that in the LSUN-bedroom dataset, the results of MemoPainter and iGM are gloomy and only locally colorized. Similarly, a mass of inharmonious yellow appears in the result of Zhang *et al.* [21]. Besides, rationality is another key point, which is especially important in some images such as human portrait. Compared with the unnatural overall yellow or red in the baby of previous methods, JGM provides a more natural result. Furthermore, benefiting from the constraint of gradient information, JGM can achieve high-quality colorization results while avoiding undesirable phenomena such as color pollution and overflow.

In general, JGM attains high PSNR and SSIM values for the colorization of a scene that is either simple or complex, which indicates that it has sound coloring effects and robustness. Besides, JGM has pleasure visual effects that can meet the colorization needs in various situations. More colorization results of JGM are provided in **Supplementary Materials**.

### C. Naturalness of Colorization Results

Generally speaking, the purpose of image colorization is to visually present a natural and reasonable effect, rather than recover the ground-truth precisely. Therefore, we conduct a user study to quantify the naturalness comparison of colorized images. In the experiment, the colorized images synthesized by different methods are presented to participants, and then they are asked to choose the most natural candidate images.

More specifically, we show the 100 observers the results and ask them to evaluate the naturalness of images based on the three indicators, including Saturability, Semantic Correctness and Edge Keeping. Among the three indicators, Saturability is used to reflect the vividness of the picture, Semantic correctness is an examination of reasonable colorization of method and Edge Keeping can describe the degree of color overflow and edges blur. After these indicators are evaluated, the overall naturalness is defined as the equally weighted sum of the three values.

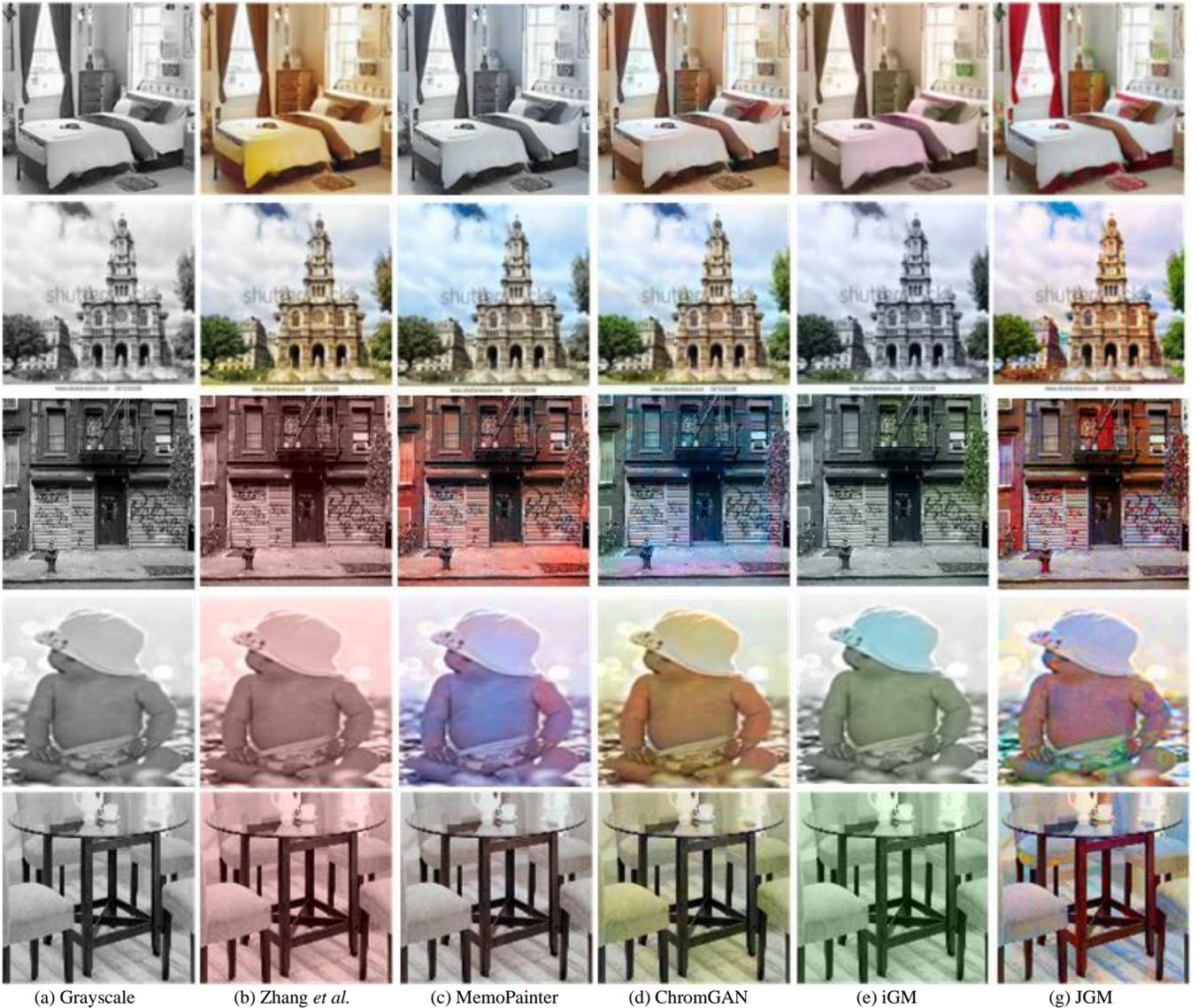

(a) Grayscale　　　(b) Zhang *et al.*　　　(c) MemoPainter　　　(d) ChromGAN　　　(e) iGM　　　(g) JGM

**Fig. 5.** Visual comparisons with the state-of-the-arts. From left to right: Grayscale, Zhang *et al.*, MemoPainter, ChromaGAN, iGM and JGM. Our method with gradient domain and high-dimensional can predict pleasing colors visually.

There are 100 participants in the experiment and 1,000 votes in total. We record their selection results and calculate the percentage of the four criteria for each image. The results are recorded in Table 2. To facilitate observation, we draw the error bar figures for comparing the images generated by different approaches as in Fig. 6.

The naturalness of iGM, Zhang *et al.*, ChromaGAN and MemoPainter is 88.96%, 88.84%, 90.44% and 89.94%, respectively. This phenomenon indicates that all the methods can achieve the goal of global naturalness but being less sensitive to local colorization. For instance, in the result of MemoPainter, there is a large area of unreasonable blue on the bedroom wall, and in Zhang *et al.* [21], strange green appears on the church ground.

In addition, other colorization methods face great challenges in terms of color saturation and semantic correctness, all of which have achieved low values. As can be seen from the results that the iGM effects are dim, Zhang *et al.* [21] and ChromGAN appear desaturated yellow, MemoPainter cannot achieve spatially consistent colorization.

It is commendable that JGM reaches the best results of 93.97%, 92.89% and 92.66% on the three indicators, indicating high saturation, contrast and semantic accuracy, without edge overflow. It can achieve distinct colors on different objects. The model with gradient information helps to achieve the highest performance and produce convincing colorizations while retaining global color consistency. In summary, it is self-evident that the results obtained by JGM have a better perception of the object color and generate the highest naturalness effect, which can be considered as the real color by subjects.

Some colorization results are zoomed in Fig. 7. In the first image, JGM can separate the small and dense complex patterns on the bed from the surroundings. In the second image, each item on the table is given a different color instead of the same overall color. At the same time, we enlarge the part of the spire and the wall of churches, and it is difficult to observe any spillover phenomenon in the results of JGM. As seen in the third image, even the leaves with twisty edges are colored accurately. In summary, due to the addition of gradient domain information, JGM exhibits amazing colorization effects, which realizes the semantic correctness and the accurate color range.

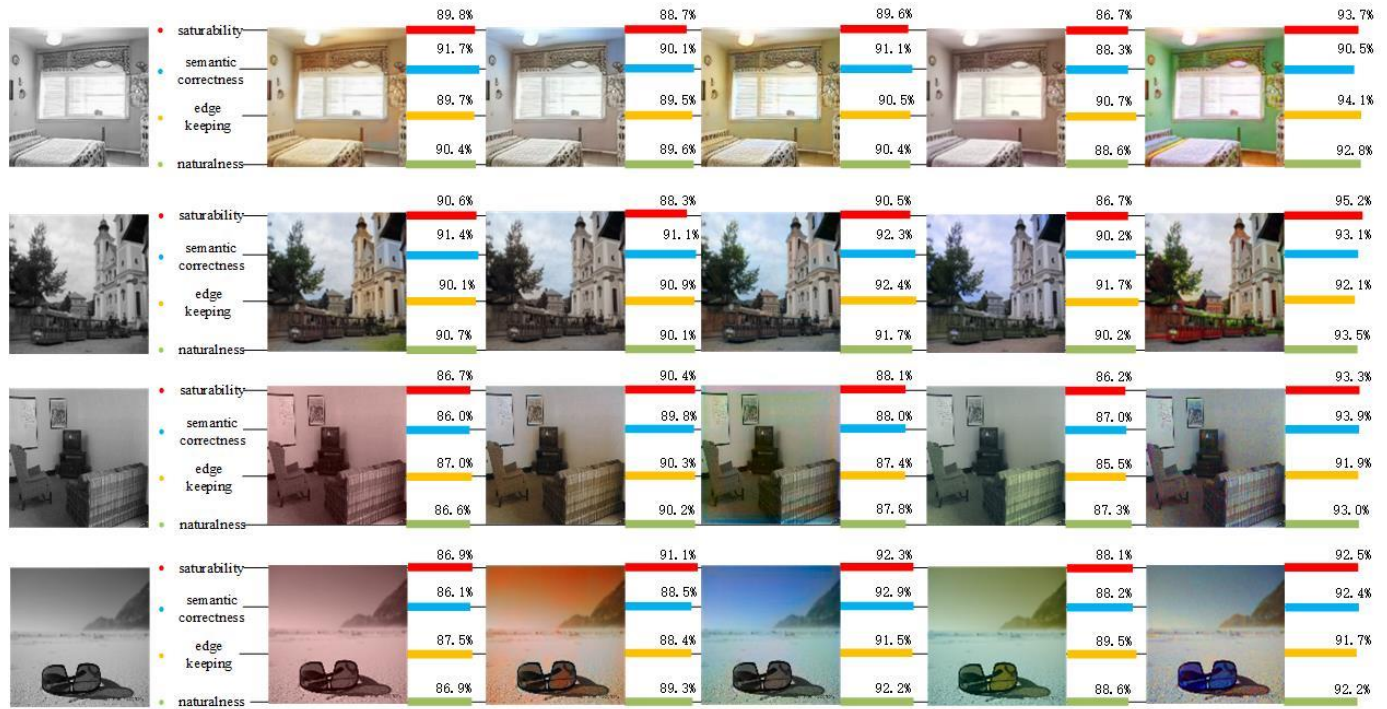

(a) Grayscale　　(b) Zhang *et al.*　　(c) MemoPainter　　(d) ChromGAN　　(e) iGM　　(g) JGM

**Fig. 6.** The exemplar of the naturalness of the state-of-the-art automatic colorization methods includes three aspects: Saturability, Semantic Correctness and Edge Keeping. JGM can produce more natural results.

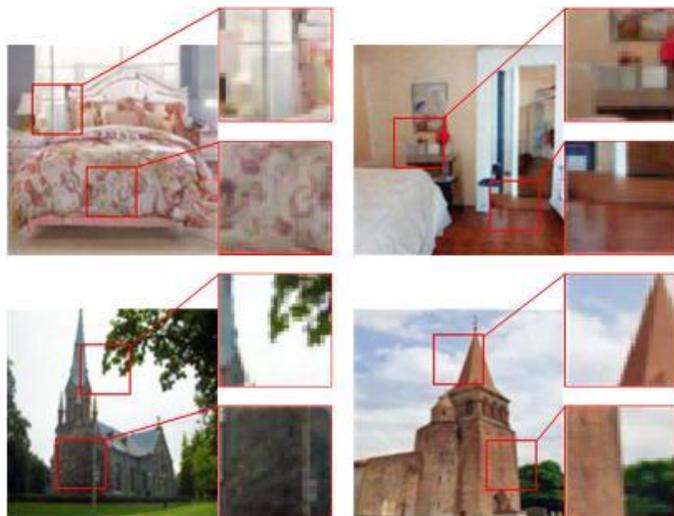

**Fig. 7.** The zoomed version of colorization results in JGM. It can be seen that the colorized images by JGM are showing high naturalness and contrast.

TABLE II
COMPARISON OF NATURALNESS BETWEEN JGM AND OTHER STATE-OF-THE-ARTS. THE MEAN VALUE OF EACH CRITERION IS RECORDED.

| Algorithms | Saturability (%) | Semantic Correctness (%) | Edge Keeping (%) | Naturalness (%) |
|---|---|---|---|---|
| Zhang *et al.* | 88.48 | 89.02 | 89.02 | 88.84 |
| MemoPainter | 90.02 | 90.12 | 89.68 | 89.94 |
| ChromaGAN | 90.30 | 90.95 | 90.06 | 90.44 |
| iGM-6C | 87.38 | 88.47 | 90.04 | 88.96 |
| JGM | **93.97** | **92.89** | **92.66** | **93.17** |

### D. Diversity of Colorization Results

Diverse colorization aims to generate different colorized images rather than restore the original color, which is often achieved via GANs [41] or VAE [42]. Due to the joint high-dimensional intensity and gradient of image information, the proposed JGM can improve the original performance of the generative model, generating multiple feasible colorizations.

As shown in Fig. 8, the diverse colorizations have reasonable effects for indoor and outdoor scenes. We observe different tones and background colors for indoor scenes and different rivers, sky for outdoor scenes. The improved diversified performances of JGM are presented as follows:

***Distinct Styles:*** In general, each image has its only style, and all the images are different in both global color contrast and chrominance. JGM reliably produces various distinct styles, such as red, blue, and yellow tones. Compared with other diversified colorization methods that appear chaotic style, JGM can bring better visual impressions.

***Uniform and No-Overflow:*** With the characteristic of intensity-gradient domain information, JGM shows more spatially coherent. The consistent overall tone of the image and no improper color pollution or overflow make our image more vivid.

*High Saturation:* JGM reduces the unnatural effect of brightness reduction and presents a rich and colorful style while maintaining high saturation and dynamic effect.

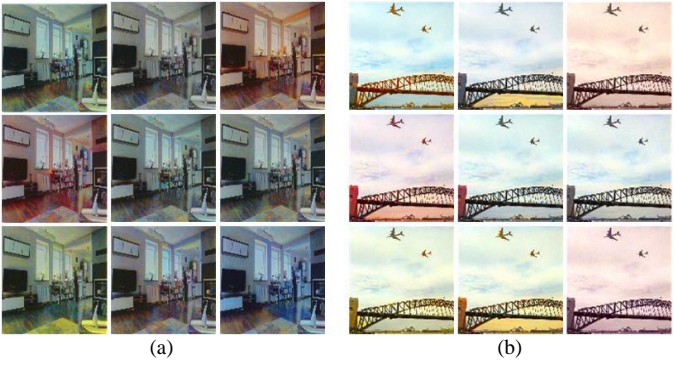

**Fig. 8.** Diversified colorization effects of the proposed JGM.

## V. DISCUSSIONS

As the joint learning in intensity-gradient domain and intensity-gradient fidelity constraint are two key factors to success of the algorithm, here we discuss the importance and superiority of high-dimensionality learning and constraints in gradient domain.

### A. Comparison of Joint/Separate Learning

In the algorithm of the JGM-divide, two networks $s_{\theta_1}(x)$ and $s_{\theta_2}(\nabla x)$ are trained in the image domain and gradient domain separately. Then, using the annealed Langevin dynamics with the prior information, image domain and gradient domain samples can be obtained from these trained networks.

Since the use of two separate models, both the image domain $x$ and the gradient domain $\nabla x$ need to be taken into consideration during the iterative colorization process. Thus, the modified annealed Langevin dynamics combined with DC flow can be rewritten as follows:

$$x_{t+1} \leftarrow x_t + \frac{\alpha_i}{2}[s_{\theta_1}(x_t) - \lambda_1 DC(x_t)] + \sqrt{\alpha_i} z_1$$
$$\nabla x_{t+1} \leftarrow \nabla x_t + \frac{\alpha_i}{2}[s_{\theta_2}(\nabla x_t) - \lambda_2 DC(\nabla x_t)] + \sqrt{\alpha_i} z_2$$
(17)

where the $DC(x) = (Fx - y)$ is the data-consistency of image domain, and $DC(\nabla x) = (F\nabla x - \nabla y)$ is the data-consistency of gradient domain.

Compared with the standard JGM, although JGM-divide can alleviate the issues of uneven colorization and insufficient contrast to some extent, it needs to train two prior information of gradient domain and image domain separately. Thus in turn, it increases the training complexity and extends the computational cost for colorization. More importantly, considering the lack of data samples in areas with low data density, DSM may not have enough evidence to accurately estimate the scoring function and the complexity of multiple models. This observation prompts us to improve performance by sampling in a high-dimensional embedding space. It can be seen from Fig. 9 that sampling in the high-dimensional embedding space can produce better results, and the contrast is higher. In brief, the visual effect and rationality of the standard JGM are better than JGM-divide.

TABLE III
COLORIZATION COMPARISON OF SEPARATE AND JOINT LEARNING UNDER GRADIENT CONSTRAINTS, IN TERMS OF PSNR AND SSIM PERFORMANCE.

| Algorithms | Church | Bedroom | COCO-stuff | ImageNet |
|---|---|---|---|---|
| JGM-divide | 23.26/**0.9269** | 22.23/0.9059 | 19.45/**0.8477** | 15.13/**0.8227** |
| JGM | **23.76**/0.8931 | **29.46/0.9365** | **20.72**/0.7254 | **18.71**/0.6821 |

### B. Comparison of with/without Gradient Data-Fidelity

To testify the influence of data-consistency constraints, we try to add gradient information without the relevant constraints to the trained model. Due to the influence of the generative model, if there is no added data-fidelity constraint of the gradient domain, the degree of freedom of image coloring will be abnormally high. It may cause the phenomenon of color overflow and unreasonable results. We provide a few failure cases in Fig. 9 (c). It is believed that incorrect gradient information will lead to low saturation and a single color of the colorization result. Although the gradient information in high-dimensional prior learning is used to improve the effect, it is not very satisfactory because of the lack of sufficient data-fidelity constraints. Therefore, enforcing more appropriate constraints at the iterative colorization stage is necessary.

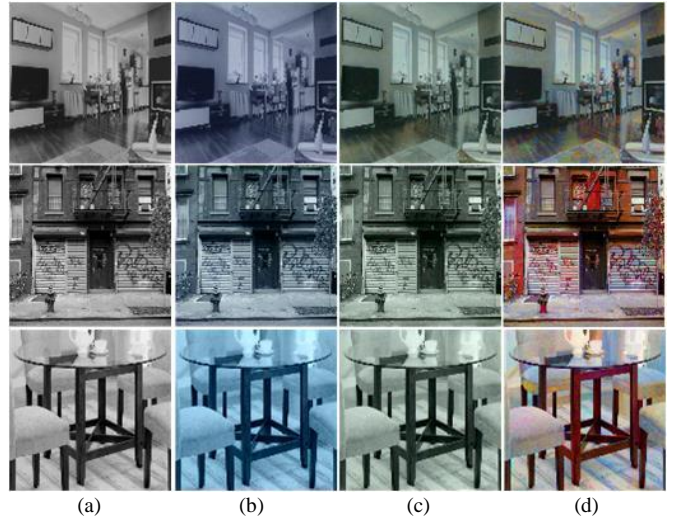

**Fig. 9.** Colorization visualization results of the JGM with different variants. (a) Grayscale, (b) JGM-divide, (c) JGM without data-fidelity constraint in gradient domain and (d) JGM.

### C. Colorization without Known Forward Operator F

In some practical scenarios, the grayscale image $y$ without known forward model $F$ is observed. In this circumstance, the task of "blind" colorization is more challenging.

At present, two prevailing processing methods of forming $F$ are:

$$Fx = (x_r + x_g + x_b)/3 \quad (18)$$

and

$$Fx = (0.3x_r + 0.59x_g + 0.11x_b) \quad (19)$$

In the following, we use a set of old historical photos to test the "blind" colorization result of JGM under these two popular forward operators $F$. Some of the results are shown in Fig. 10. It can be seen that, by assuming two different forward operators, the coloring effect has some deviations in details such as brightness and saturation. Due to the robustness of the network-based prior information and the flexibility of the prior information to different forward models, both effects are reasonable.

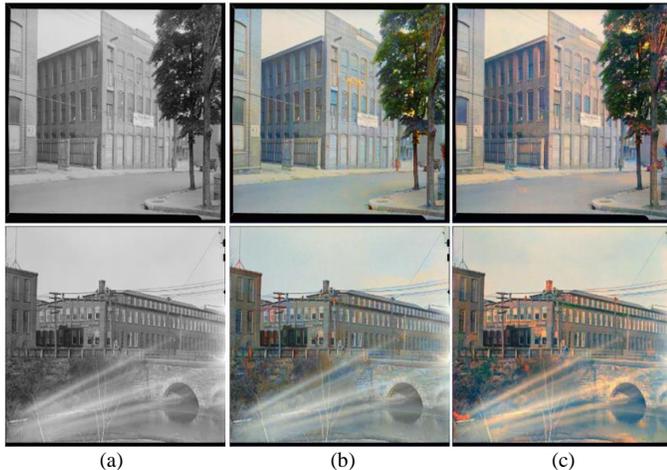

(a)            (b)            (c)

**Fig. 10.** "Blind" colorization results of historical black and white photographs from the early 20th century. These pictures [60] are *Textile Mill, June 1937* and *Scott's Run, Mar. 1937* taken from the US National Archives (Public Domain). (a) The observed image. (b) JGM with Eq (18). (c) JGM with Eq (19).

## VI. CONCLUSIONS

In this work, an iterative generative model for automatic colorization with two major characteristics was proposed, namely, prior learning in joint intensity-gradient domain and iterative colorization in joint intensity-gradient field. More specifically, generative modeling in high-dimensional space has addressed major limitations of existing methods and provided diverse probable colorizations. Additionally, enforcing data-fidelity in joint intensity-gradient field produced substantially improved results, as validated by extensive comparisons with state-of-the-art methods. The proposed JGM only involves a couple of insensitive parameters, which are fixed in all experiments. In the forthcoming study, we hope to apply these critical factors in wavelet domain, etc. Besides, extending the core idea used in colorization to hyperspectral imaging is also a promising direction [61].

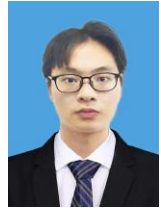

**Kai Hong** is studying for the M. S. degree in the School of Information Engineering, Nanchang University. His current research interests include sparse representation theory, deep learning and its application in image processing and colorization.

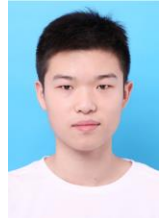

**Jin Li** was born in Jiangxi, China, in 2000. He is pursuing a bachelor's degree in the School of Information Engineering, Nanchang University, Nanchang, China. His current research interests include deep learning and its application in image processing and colorization

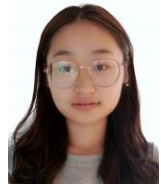

**Wanyun Li** was born in Sichuan, China, in 2000. She is pursuing a bachelor's degree in the School of Information Engineering, Nanchang University, Nanchang, China. Her current research interests include deep learning and its application in image processing and colorization

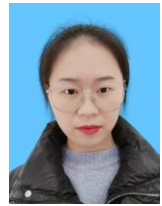

**Cailian Yang** is studying for the M. S. degree in the School of Information Engineering, Nanchang University. Her current research interests include sparse representation theory, deep learning and its application in image processing and colorization.

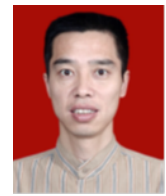

**Minghui Zhang** was born in1963. He received the M.S. degree from Chongqing University, Chongqing, in 1990, majored in Biomedical engineering. Professor Zhang's research interest includes MRI reconstruction, image compression and restoration, pattern recognition, etc.

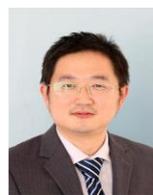

**Yuhao Wang** (SM'14) received the Ph.D. degree from Wuhan University. He was a Visiting Professor at University of Calgary in 2008, and also with China National Mobile Communication Research Laboratory, Southeast University from 2010 to 2011. He is currently a Professor in School of Information Engineering, Nanchang University. His current research interests include wideband wireless communication and radar sensing fusion system, channel measurement and modeling, smart sensor, image and video processing, and machine learning, visible light communication.

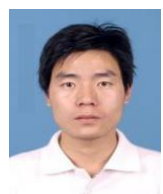

**Qiegen Liu** (SM'19) received the Ph.D. degree in Biomedical Engineering from Shanghai Jiao Tong University. Since 2012, he has been with School of Information Engineering, Nanchang University, Nanchang, China, where he is currently an Associate Professor. During 2015-2017, he is also a postdoc in UIUC and University of Calgary. His current research interest is sparse representations, deep learning and their applications in image processing, computer vision and MRI reconstruction.